\documentclass[10pt,twocolumn,letterpaper]{article}

\usepackage[pagenumbers]{cvpr} 

\definecolor{cvprblue}{rgb}{0.21,0.49,0.74}
\usepackage[pagebackref,breaklinks,colorlinks,allcolors=cvprblue]{hyperref}

\usepackage{multirow}
\def\mymodel{ReVSeg}

\usepackage{graphicx}
\usepackage{resizegather}
\usepackage{colortbl}
\usepackage{amssymb}
\usepackage{dsfont}
\usepackage[ruled,linesnumbered]{algorithm2e}

\title{ReVSeg: Incentivizing the Reasoning Chain for Video Segmentation \\ with Reinforcement Learning}

\author{
Yifan Li\textsuperscript{1,2} \quad Yingda Yin\textsuperscript{3}\thanks{Corresponding authors.} \thanks{Co-project leader} \quad Lingting Zhu\textsuperscript{3\dag} \quad Weikai Chen\textsuperscript{3} \quad Shengju Qian\textsuperscript{3} \\ Xin Wang\textsuperscript{3} \quad Yanwei Fu\textsuperscript{1,2*}\\
\normalsize \textsuperscript{1}Fudan University\quad
\textsuperscript{2}Shanghai Innovation Institute\quad
\textsuperscript{3}LIGHTSPEED
}

\begin{document}

\twocolumn[{%
\renewcommand\twocolumn[1][]{#1}%
\maketitle
\vspace{-24pt}
\begin{center}
    \centering
    \captionsetup{type=figure}
    \includegraphics[width=0.95\linewidth]{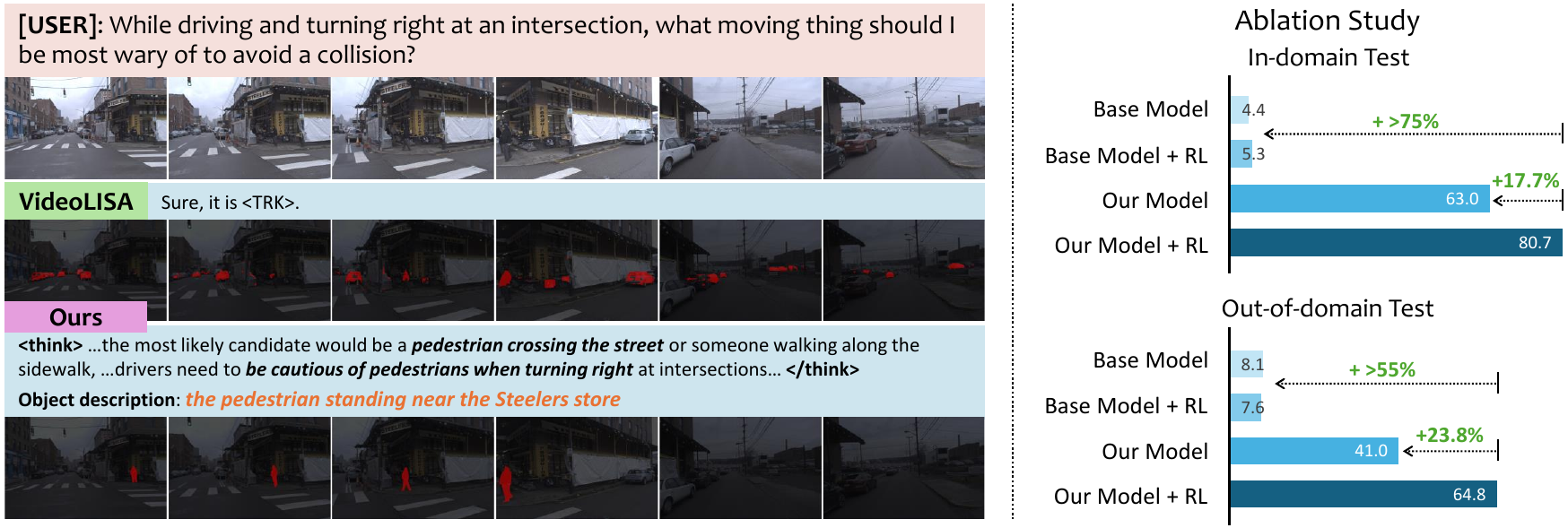} 
    \vspace{-6pt}
    \captionof{figure}{
    (Left) Through an explicit reasoning chain, our \mymodel\ tackles reasoning-focused video object segmentation and accurately grounds objects referenced by complex, abstract real-world queries.
    (Right) While the base model and its RL variant struggle on the task, our method achieves strong performance, with RL post-training yielding a further substantial boost.
    We report the $\mathcal{J}\&\mathcal{F}$ metric on Ref-DAVIS17 (in-domain) and ReasonVOS (out-of-domain) datasets in the chart.
    }
    \label{fig:teaser}
\end{center}%
}]

{
\renewcommand{\thefootnote}{}%
\footnotetext{\textsuperscript{*} Corresponding authors.}
\footnotetext{\textsuperscript{\dag} Project leads.}%
}

\begin{abstract}
Reasoning-centric video object segmentation is an inherently complex task: the query often refers to dynamics, causality, and temporal interactions, rather than static appearances. Yet existing solutions generally collapse these factors into simplified reasoning with latent embeddings, rendering the reasoning chain opaque and essentially intractable.
We therefore adopt an explicit decomposition perspective and introduce \mymodel, which executes reasoning as sequential decisions in the native interface of pretrained vision language models (VLMs). 
Rather than folding all reasoning into a single-step prediction, \mymodel\ executes three explicit operations -- semantics interpretation, temporal evidence selection, and spatial grounding -- aligning pretrained capabilities.
We further employ reinforcement learning to optimize the multi-step reasoning chain, enabling the model to self-refine its decision quality from outcome-driven signals. 
Experimental results demonstrate that \mymodel\ attains state-of-the-art performances on standard video object segmentation benchmarks and yields interpretable reasoning trajectories. \href{https://clementine24.github.io/ReVSeg/}{Project Page}.
\end{abstract}    
\section{Introduction}
\label{sec:intro}

Human interpretation of videos relies on recognizing how events unfold, why actions occur, and when key moments matter.
Traditional video object segmentation (VOS) methods, however, primarily exploit appearance-only or category-level cues~\cite{wang2021end,wu2021seqformer,botach2022end,seo2020urvos,wu2022language}.
Reasoning VOS elevates the task: the model must parse abstract, under-specified instructions, draw on commonsense and causal reasoning, integrate temporal dynamics with semantic knowledge to identify the target (e.g., ``the runner most likely to win'' or ``the object causing the accident'').

While recent vision-language models (VLMs) have shown promising capabilities for reasoning segmentation~\cite{lai2024lisa,yan2024visa,bai2024one}, current systems predominantly model reasoning VOS as a single-step latent prediction: inserting a special token (e.g., $<$SEG$>$) and decoding it directly into mask outputs~\cite{bai2024one,wang2025object,yan2024visa,gong2025devil,lin2025glus,wei2025hyperseg,wei2025instructseg,zheng2025villa}.
This collapses the multi-step reasoning process — interpreting abstract instructions, identifying candidates, and spatial-temporal grounding — into a simple conclusion as well as opaque embeddings.
Such compactness comes at a great cost: interpretability is bounded, distribution shift arises from forcing VLMs into non-native output spaces, and substantial data is required for supervised fine-tuning.

These challenges reflect a deep insight: effective video reasoning unfolds through \textit{a sequence of deliberate choices}, not a single latent inference.
A model must determine \textit{where} to focus, \textit{when} to attend, and \textit{which} entity the query refers to.
Motivated by this principle, we introduce \mymodel, which eliminates the use of latent segmentation tokens and instead {re}formulates {re}asoning video segmentation as an explicit sequence of reasoning chain.
In particular, \mymodel\ decomposes reasoning VOS into three actions aligned with VLM-native capabilities: video understanding (interpreting the query and assessing scene dynamics), temporal grounding (identifying key frames or intervals pertinent to the query), and spatial grounding (localizing the target objects within selected frames). 
\mymodel\ orchestrates these primitives through multi-turn actions with a single VLM, ensuring the semantic context established in early reasoning steps seamlessly propagates to downstream localization.
Executing reasoning through these native interfaces preserves pretraining alignment, avoids heavy re-training, and transforms an otherwise entangled problem into a structured procedure.

Once the reasoning chain is made explicit, the central challenge becomes how to optimize the chain itself. 
Supervised learning provides little leverage, as the correctness of intermediate decisions is hardly annotated. 
We therefore adopt \textit{reinforcement learning} (RL)~\cite{shao2024deepseekmath,guo2025deepseek}, viewing reasoning VOS as a behavioral policy that should be rewarded only when its full decision trajectory leads to a correct segmentation outcome~\cite{li2025videochat}.
Yet, na\"ively applying RL encounters obstacles: reasoning VOS offers extremely sparse success signals, and requires coordination across video understanding, temporal selection, and spatial localization. 
To address these challenges, we introduce reasoning-aligned rewards which inject signals at critical decision points, ensuring that RL enhances the model’s reasoning behavior itself.
Together with decomposition, RL transforms the task from an opaque latent regression into a structured, optimizable reasoning policy: decomposition isolates what the model must decide while RL determines how those decisions should cooperate.
This synergy yields a reasoning policy that is both internally consistent and empirically strong.

Experiments show that \mymodel\ delivers the state-of-the-art (SOTA)
performances on multiple standard VOS benchmarks, outperforming both latent-based VLMs \cite{lai2024lisa,zhu2023tracking,yan2024visa,gong2025devil,lin2025glus,wei2025hyperseg,wang2025object,zheng2025villa,wei2025instructseg} and the training-free explicit VLM \cite{CoTRVS},
with auditable reasoning traces. 
As shown in Figure~\ref{fig:teaser}, controlled ablations demonstrate that without our decomposed reasoning chain, both the base VLM and its RL post-trained variant tend to fail on the complex reasoning VOS task. In contrast, our framework attains strong performance even before RL, and the RL post-training further delivers a substantial improvement.
In summary, our contributions are as follows:
\begin{itemize}
    \item \textit{A principled decomposition of reasoning VOS.}
    We introduce \mymodel, which \textbf{re}formulates \textbf{re}asoning \textbf{v}ideo \textbf{seg}mentation as an explicit multi-step reasoning chain built from native VLM primitives.  
    \item \textit{An RL framework that optimizes the reasoning chain itself.}
    We develop a novel RL-based formulation that directly optimizes the reasoning chain, allowing the model to refine decision quality without requiring dense supervision.
    \item We set the new state of the art on standard VOS benchmarks while providing interpretable reasoning traces.
\end{itemize}

\section{Related Works}
\label{sec:related work}
\begin{figure*}[]
    \centering
    \includegraphics[width=0.95\textwidth]{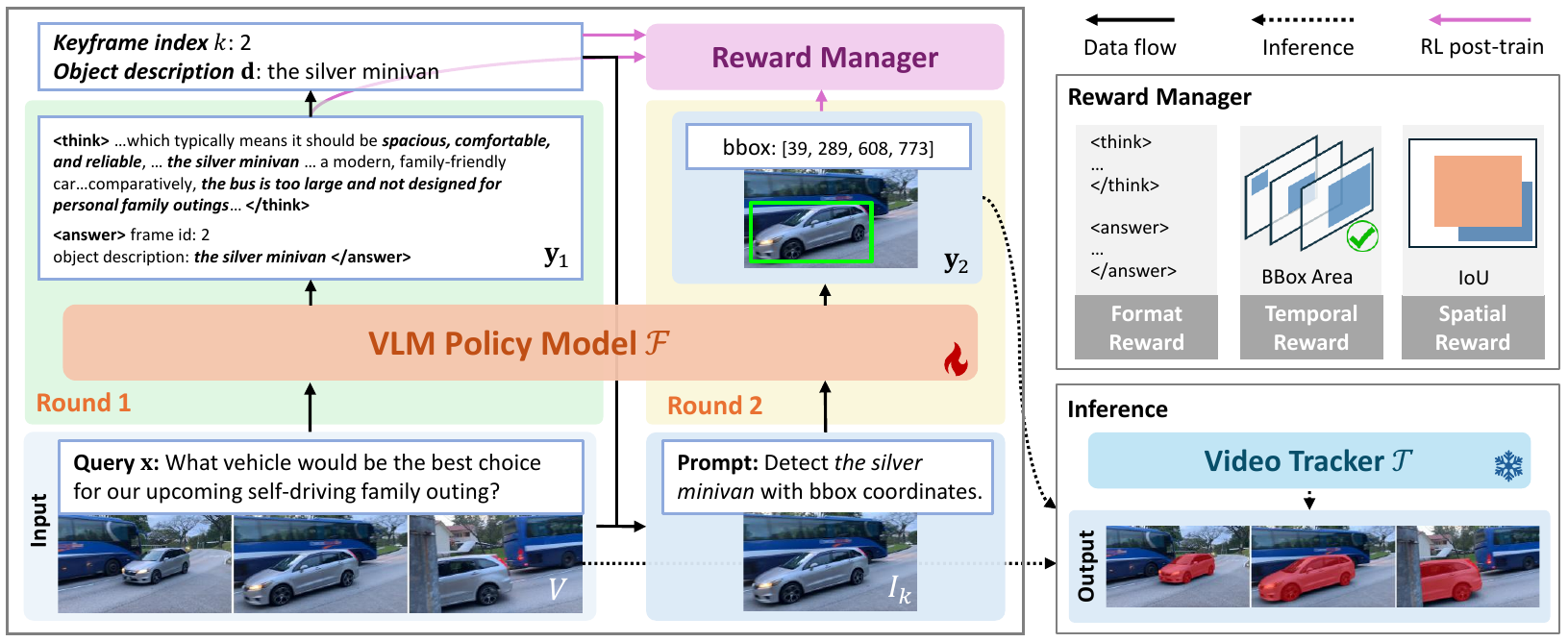} 
     \vspace{-0.1in}
    \caption{Overview of \mymodel. The model runs a two-turn reasoning chain over the input video and query. Round one analyzes the scene and selects an informative keyframe with a concise object description. Round two grounds the target on that keyframe by predicting a bounding box. The keyframe-bbox pair conditions a video tracker to produce full segmentation sequence. A reward manager provides concise signals to post-train the VLM via  reinforcement learning, improving keyframe selection, grounding accuracy, and overall robustness.  \label{fig:framework} }
    \vspace{-0.15in}
   
\end{figure*}

\subsection{Reasoning Video Object Segmentation}
Fine-grained text-guided VOS has advanced through specialized segmentation architectures and referring-video pipelines~\cite{wu2021seqformer,botach2022end,ding2023mevis,seo2020urvos}. For the emerging and more complex setting of reasoning VOS, recent methods fine-tune VLMs to emit implicit mask tokens~\cite{bai2024one,yan2024visa,gong2025devil,lin2025glus,wei2025hyperseg,wei2025instructseg,zheng2025villa}, often paired with strong decoders such as SAM/SAM2~\cite{kirillov2023segment,ravi2024sam}. Yet robust reasoning in complex videos—multiple similar objects, occlusion, fast motion, and long-range context—remains challenging. Explicit textual reasoning via CoT has begun to appear, \eg, the training-free framework CoT-RVS~\cite{CoTRVS}, which adopts two independent VLM systems for frame-by-frame segmentation bridged via text.
While effective, such designs are bounded by the separated information flow and module interoperability and is constrained to be training-free.
In this work, we formulate reasoning VOS as an explicit reasoning chain within a unified VLM, and employ reinforcement learning for self-improvement.

\subsection{VLMs with Reinforcement Learning}
Reinforcement learning has emerged as a practical route to strengthen the reasoning capabilities of large models when supervised annotated data are scarce. Test-time scaling via Chain-of-Thought~\cite{wei2022chain} improves VLM reasoning, while rewards further refine solution quality. Notably, Group Relative Policy Optimization (GRPO)~\cite{shao2024deepseekmath} enables efficient critic-free updates and strong gains with modest training budgets, as demonstrated by DeepSeek-Zero~\cite{guo2025deepseek}. Recent efforts~\cite{hong2025glm,wu2025mmsearch,wang2025skywork,team2025kimi,meng2025mm} leverage VLMs for high-level reasoning, and reasoning image grounding~\cite{liu2025seg,zheng2025deepeyes,liu2025visionreasoner,sarch2025grounded,wu2025vtool} has been actively studied. However, in the video modality, most work~\cite{feng2025video,li2025videochat,chen2024longvila} still targets high-level understanding; fine-grained spatio-temporal grounding remains limited due to task complexity. Our results show that reinforcement learning can effectively boost reasoning for video grounding.

\section{Method}
\label{sec:method}

\subsection{Overview}
\noindent\textbf{Task Formulation.} 
Given a natural language query $\mathbf{x}$ and a video sequence with $T$ frames $V=\{I_t\}_{t=1}^T \in \mathbb{R}^{T \times H \times W \times 3}$, where $H$ and $W$ denote the height and width of each frame, the goal of reasoning VOS is to segment the query-referred objects throughout the video. The model predicts a sequence of binary masks $M=\{m_t\}_{t=1}^{T} \in \mathbb{R}^{T \times H \times W}$, where $m_t \in \{0,1\}^{H \times W}$ represents the foreground region in frame $t$. Let $M^*=\{m_t^*\}_{t=1}^{T}$ denote the ground-truth masks. The task is to learn a mapping
\begin{equation}
    (V,\mathbf{x}) \rightarrow M,
\end{equation}
that maximizes agreement with $M^*$. This mapping must correctly resolve the semantic correspondence specified in query and maintain temporal consistency across the video.

{\noindent\textbf{Pipeline Overview.}}
\mymodel\ formulates reasoning-centric video object segmentation as an explicit sequence of reasoning steps. The complex task is decomposed into three VLM-native capabilities -- video understanding, temporal grounding, and spatial grounding -- and executed through multi-turn dialogue with a single VLM.

As illustrated in \cref{fig:framework}, the \textbf{first-round dialogue} takes the video and query $(V,\mathbf{x})$ as input. The VLM performs video understanding and temporal grounding: it interprets the abstract query, analyzes scene dynamics, produces a concise spatial description of the target, and identifies the keyframe that best captures the entity of interest.
Within this round, the raw, cluttered video sequence and the vague high-level query (e.g., ``the best choice for the family outing'') are distilled into a well-specified keyframe and a concrete textual description (e.g., ``the silver minivan''). These intermediate outputs encapsulate the VLM’s commonsense and causal reasoning, effectively transforming a complex video–text reasoning problem into a substantially simpler image-level segmentation task.

Given these intermediate results, the \textbf{second-round dialogue} processes the selected keyframe and the concrete object description. The VLM then performs spatial grounding and outputs a tight bounding box for the specified object, completing the two-round reasoning chain.
With the localized keyframe target, an off-the-shelf video tracker (\eg., SAM2) is readily applied to produce the final video mask predictions.

Because both rounds occur within a single VLM, the semantic context established during early reasoning is seamlessly propagated to subsequent steps, ensuring consistency throughout the chain.

Finally, we optimize the reasoning process through reinforcement learning. Rather than rewarding only the final outcome of the complex task, the decomposition of the pipeline and the availability of meaningful intermediate products allow for richer, reasoning-aligned reward signals that better guide policy improvement.

\subsection{Decomposed Generation with Reasoning Chain}
\label{sec:decomposed generation}

Directly producing spatio-temporal grounding from video input remains challenging for current VLMs~\cite{bai2025qwen2,wang2025internvl3_5,wang2024qwen2,hong2025glm,team2025kimi,team2025gemma}. Owing to the limited availability of high-quality annotated datasets, these models have not been extensively pretrained for VOS.
Even with test-time scaling strategies -- such as Chain-of-Thought prompting~\cite{wei2022chain}, multi-rollout sampling with self-consistency~\cite{wang2022self}, or search-based inference methods including Tree Search and Beam Search~\cite{yao2023tree,vijayakumar2016diverse} -- VLMs still struggle to generate reliable spatio-temporal grounding results directly from raw video.

These limitations highlight a key observation that complex video reasoning inherently unfolds as a sequence of interdependent reasoning actions, rather than a single-step prediction. This insight motivates us to decompose the reasoning-based VOS task into a set of primitive capabilities, orchestrated through a multi-turn reasoning chain, as described below.

\noindent\textbf{First Round Rollout.} 
In the first reasoning round, the VLM $\mathcal{F}$ receives a video–query pair $(V, \mathbf{x})$ and generates a text response $\mathbf{y_1}$ under an instruction-guided prompt.
During this step, the model interprets the user query, analyzes the video content, infers the target entity, identifies its temporal occurrence within the sequence, and produces a concise textual description of the target object grounded in a keyframe.
The output of this reasoning round can be formalized as:
\begin{equation}
\mathbf{y_1} \sim \mathcal{F}(\cdot \mid V, \mathbf{x}).
\end{equation}

A parser $\mathcal{G}$ then processes the structured response $\mathbf{y_1}$, extracting: the keyframe index $k\in\{0,1,\cdots,T-1\}$, a concise spatial description $\mathbf{d}$ of target objects, and a status flag $S_1\in\{\texttt{succ}, \texttt{fail}\}$ indicating the extraction success:
\begin{equation}
    \mathcal{G}(\mathbf{y_1})=\left\{
    \begin{aligned}
    &(S_1, k, \mathbf{d}), &S_1=\texttt{succ}\\
    &(S_1,\texttt{null},\texttt{null}),&S_1=\texttt{fail}.
    \end{aligned}
    \right .
\end{equation}
Based on the selected keyframe index $k$, the corresponding frame $I_k$ is retrieved from the video sequence $V$.

\noindent\textbf{Second Round Rollout.}
If the status flag from the previous round is $S_1=\texttt{succ}$, the rollout generation proceeds to the second stage. In this round, the VLM $\mathcal{F}$ receives the selected keyframe $I_k$ and the concise object description $\mathbf{d}$, together with the generation history $(V,\textbf{x},\mathbf{y_1})$ from the first round.
Conditioned on this context and an instruction-guided prompt, the model generates a text response $\mathbf{y_2}$.
At this stage, $\mathcal{F}$ is responsible for spatial grounding: it localizes the target object in keyframe $I_k$ using both the visual input and the accumulated dialogue history. The output $\mathbf{y_2}$ follows a structured format, formalized as:
\begin{equation}
    \mathbf{y_2}\sim\mathcal{F}(\cdot|V,\mathbf{x},\mathbf{y_1},I_k,\mathbf{d}).
\end{equation}
Similarly, the parser $\mathcal{G}$ processes the structured response $\mathbf{y_2}$, extracting a bounding box $B_k\in\mathbb{R}^4_{\geq 0}$ and a new status flag $S_2$:
\begin{equation}
    \mathcal{G}(\mathbf{y_2})=\left\{
    \begin{aligned}
    &(S_2, B_k), &S_2=\texttt{succ}\\
    &(S_2,\texttt{null}),&S_2=\texttt{fail}.
    \end{aligned}
    \right .
\end{equation}
The final rollout output $o$ is obtained by concatenating the two-stage responses:
\begin{equation}
    o = \mathbf{y_1}\oplus \mathbf{y_2}.
\end{equation}
This two-round decomposition cleanly separates video understanding and temporal selection from spatial localization, facilitates well-defined task at each stage, and provides a stable, modular interface for the RL optimization.

\subsection{Reasoning VOS with GRPO}
\label{sec:rl}

Reasoning capability is the key factor determining the upper bound of reasoning-based VOS. However, the supervised reasoning data required by Supervised Fine-tuning (SFT) is scarce and costly. Inspired by outcome-driven training in recent reasoning systems~\cite{guo2025deepseek,shao2024deepseekmath}, we employ reinforcement learning to enable the policy model $\mathcal{F}$ to self-improve under task-specific rewards, thereby enhancing its reasoning ability.

\noindent\textbf{Reward Modeling.} 
Rewards shape the optimization dynamics and are therefore crucial in RL. Following the minimalist design philosophy of DeepSeek-R1-Zero~\cite{guo2025deepseek}, we adopt a rule-based reward system consisting of three components:
\begin{itemize}
    \item \textit{Format Reward} $r_f$: Correct output formatting is essential for interacting with the environment. The model must place its reasoning process between \texttt{<think>} and \texttt{</think>} tags and the final answer between \texttt{<answer>} and \texttt{</answer>} tags. In addition, the first-turn output $\mathbf{y_1}$ must include the keyframe index $k$ and object description $\mathbf{d}$ in JSON format, while the second-turn output $\mathbf{y_2}$ must provide the bounding box $B_k$ in JSON. Based on the degree to which the output $o$ satisfies these rules, the format reward $r_f$ is assigned a value in $[0,1]$.
    \item \textit{Temporal Reward} $r_t$: The keyframe $I_k$ selected by $\mathcal{F}$ in $\mathbf{y_1}$ critically affects subsequent spatial grounding. Beyond merely containing the target object, we encourage selecting frames where the object is clearly visible, minimally occluded, and sufficiently large. We experiment with several temporal reward choices and finally choose the normalized area of the ground-truth bounding box at $I_k$: 
    \begin{equation}
        r_t=\mathds{1}_{(||m_k^*||_1>0)}\cdot \frac{{\mathcal{S}}(m_k^*)-\min_{t}{\mathcal{S}}(m_t^*)}{\max_{t}{\mathcal{S}}(m_t^*)-\min_{t}{\mathcal{S}}(m_t^*)}.
    \end{equation}
    where $\mathcal{S}(m^*_t)$ denotes the pixel area of the ground truth bbox at frame $I_t$ and $\mathds{1}_{(\cdot)}$ denotes the indicator function. See \cref{sec:temporal rewards} for ablations.
    \item \textit{Spatial Reward} $r_s$: This reward measures final detection quality of the predicted $B_k$. Following prior works~\cite{liu2025seg,liu2025visionreasoner}, we use Intersection-over-Union (IoU) between the predicted and ground-truth boxes. If $\text{IoU} > 0.5$, the prediction is considered correct and $r_s = 1$; otherwise $r_s = 0$.
\end{itemize}

The total reward for an output $o$ combines these components with status flags $S_1$ and $S_2$ indicating whether each step succeeds:
\begin{equation}
    r = r_f + \mathds{1}_{(S_1=\texttt{succ})} r_t + \mathds{1}_{(S_1=\texttt{succ} \ \&\  S_2=\texttt{succ})} r_s.
\end{equation}

\noindent\textbf{Objective.} 
We adopt Group Relative Policy Optimization (GRPO)~\cite{shao2024deepseekmath}, a critic-free variant of Proximal Policy Optimization (PPO)~\cite{schulman2017proximal} tailored for sequence models. Given an input query $q=\{V,\mathbf{x}\}$, GRPO samples a group of $n$ candidate outputs $\{o_i\}_{i=1}^n$ from the current policy $\pi_{\text{old}}$, evaluates their rewards $\{r_i\}$, and computes a normalized within-group advantage:
\begin{equation}
    A_i = \frac{r_i - \text{mean}(r_1, \ldots, r_n)}{\text{std}(r_1, \ldots, r_n)}.
\end{equation}
Since we use on-policy updates ($\pi_{\text{old}} = \pi_\theta$) in practice, importance sampling ratios are remains at one. The policy loss, combined with KL regularization to a reference model $\pi_{\text{ref}}$, yields the final training objective:
\begin{equation}
\mathcal{L}(\theta) = 
\mathbb{E}_{q \sim P(Q),\, \{o_i\} \sim \pi_\theta(\cdot|q)}
\left[
\frac{1}{n} \sum_{i=1}^n 
\Big( A_i - \beta\,\mathbb{D}_{\text{KL}}(\pi_\theta \,\|\, \pi_{\text{ref}}) \Big)
\right].
\end{equation}

\section{Experiments}
\label{sec:experiments}

\subsection{Experiment Settings}
\label{sec: experiment settings}

\begin{table} \small 
\centering
\caption{Reasoning video object segmentation performance comparison on ReasonVOS~\cite{bai2024one} dataset.\label{tab:reasonvos} }
\begin{tabular}{llccc}
\hline
\multicolumn{2}{c}{Method}                                                                & $\mathcal{J}$ & $\mathcal{F}$ & $\mathcal{J}\&\mathcal{F}$ \\ \hline
MTTR~\cite{botach2022end}            & {\color[HTML]{656565} \tiny [CVPR\textquotesingle22]}    & 29.1          & 33.1          & 31.1                       \\
ReferFormer~\cite{wu2022language}    & {\color[HTML]{656565} \tiny [CVPR\textquotesingle22]}    & 30.2          & 35.6          & 32.9                       \\
SOC~\cite{luo2023soc}                & {\color[HTML]{656565} \tiny [NeurIPS\textquotesingle24]} & 33.3          & 38.5          & 35.9                       \\
OnlineRefer~\cite{wu2023onlinerefer} & {\color[HTML]{656565} \tiny [CVPR\textquotesingle23]}    & 34.6          & 42.9          & 38.7                       \\
SgMg~\cite{miao2023spectrum}         & {\color[HTML]{656565} \tiny [ICCV\textquotesingle23]}    & 33.7          & 38.7          & 36.2                       \\ \hline
LISA~\cite{lai2024lisa}              & {\color[HTML]{656565} \tiny [CVPR\textquotesingle24]}    & 29.1          & 33.1          & 31.1                       \\ 
VideoLISA~\cite{bai2024one}          & {\color[HTML]{656565} \tiny [NeurIPS\textquotesingle24]} & 45.1          & 49.9          & 47.5                       \\
GLUS~\cite{lin2025glus}              & {\color[HTML]{656565} \tiny [CVPR\textquotesingle25]}    & 47.5          & 52.4          & 49.9                       \\
RGA-3B~\cite{wang2025object}         & {\color[HTML]{656565} \tiny [ICCV\textquotesingle25]}    & 49.1          & 54.3          & 51.7                       \\
RGA-7B~\cite{wang2025object}         & {\color[HTML]{656565} \tiny [ICCV\textquotesingle25]}    & 51.3          & 56.0          & 53.6                       \\ \hline
CoT-RVS-online-7B~\cite{CoTRVS}                                          & {\color[HTML]{656565} \tiny [arXiv\textquotesingle25]}    & 49.5          & 54.5          & 52.0                       \\
CoT-RVS-offline-13B~\cite{CoTRVS}                                        & {\color[HTML]{656565} \tiny [arXiv\textquotesingle25]}   & 47.5          & 54.0          & 50.7                       \\
\rowcolor{gray!30}\mymodel-7B   (Ours)                                          & \multicolumn{1}{c}{\color[HTML]{656565} \tiny -}  & \textbf{61.8}          & \textbf{67.7}          & \textbf{64.8}                       \\ \hline
\end{tabular}
 \vspace{-0.1in}
\end{table}

\begin{table*}[]\footnotesize\centering
\caption{Reasoning video object segmentation performance comparison on ReVOS~\cite{yan2024visa} dataset.\label{tab:revos} }
 \vspace{-0.05in}
\begin{tabular}{llcccccccccc}
\hline
\multicolumn{2}{c}{\multirow{2}{*}{Method}} & \multicolumn{1}{c}{\multirow{2}{*}{Type}}                                                                                 & \multicolumn{3}{c}{referring}                                & \multicolumn{3}{c}{reasoning} & \multicolumn{3}{c}{overall}  \\ \cline{4-12}
                        &                            &                                                                                                       & $\mathcal{J}$ & $\mathcal{F}$ & $\mathcal{J}\&\mathcal{F}$ & $\mathcal{J}$ & $\mathcal{F}$ & $\mathcal{J}\&\mathcal{F}$ & $\mathcal{J}$ & $\mathcal{F}$ & $\mathcal{J}\&\mathcal{F}$                               \\ \hline
MTTR~\cite{botach2022end}                    & {\color[HTML]{656565} \tiny [CVPR\textquotesingle22]}                   & \multirow{4}{*}{\begin{tabular}[c]{@{}c@{}}Segmentation\\ Specialists\end{tabular}}                   & 29.8                     & 30.2                     & 30.0   & 20.4    & 21.5    & 21.0      & 25.1    & 25.9   & 25.5                                 \\
ReferFormer~\cite{wu2022language}             & {\color[HTML]{656565} \tiny [CVPR\textquotesingle22]}                   &                                                                                                       & 31.2                     & 34.3                     & 32.7   & 21.3    & 25.6    & 23.4      & 26.2    & 29.9   & 28.1                              \\
LMPM~\cite{ding2023mevis}                    & {\color[HTML]{656565} \tiny [ICCV\textquotesingle23]}                   &                                                                                                       & 29.0                     & 39.1                     & 34.1   & 13.3    & 24.3    & 18.8      & 21.2    & 31.7   & 26.4                            \\
LLaMA-VID~\cite{li2024llama} + LMPM~\cite{ding2023mevis}        & {\color[HTML]{656565} \tiny [ECCV\textquotesingle24]}                      &                                                                                                       & 29.0                     & 39.1                     & 34.1   & 12.8    & 23.7    & 18.2      & 20.9    & 31.4   & 26.1                             \\ \hline
LISA-7B~\cite{lai2024lisa}                 & {\color[HTML]{656565} \tiny [CVPR\textquotesingle24]}                   & \multirow{13}{*}{\begin{tabular}[c]{@{}c@{}}VLM-Based\\  w/ Latent \\ Tokens\end{tabular}}                          & 44.3                     & 47.1                     & 45.7   & 33.8    & 38.4    & 36.1      & 39.1    & 42.7   & 40.9                                \\
LISA-13B~\cite{lai2024lisa}                & {\color[HTML]{656565} \tiny [CVPR\textquotesingle24]}                   &                                                                                                       & 45.2                     & 47.9                     & 46.6   & 34.3    & 39.1    & 36.7      & 39.8    & 43.5   & 41.6                               \\
TrackGPT(IT)-7B~\cite{zhu2023tracking}         & {\color[HTML]{656565} \tiny [arXiv\textquotesingle24]}                  &                                                                                                       & 46.7                     & 49.7                     & 48.2   & 36.8    & 41.2    & 39.0      & 41.8    & 45.5   & 43.6                              \\
TrackGPT(IT)-13B~\cite{zhu2023tracking}        & {\color[HTML]{656565} \tiny [arXiv\textquotesingle24]}                  &                                                                                                       & 48.3                     & 50.6                     & 49.5   & 38.1    & 42.9    & 40.5      & 43.2    & 46.8   & 45.0                              \\
VISA-7B~\cite{yan2024visa}                 & {\color[HTML]{656565} \tiny [ECCV\textquotesingle24]}                   &   & 51.1                     & 54.7                     & 52.9   & 36.7    & 41.7    & 39.2      & 43.9    & 48.2   & 46.1                              \\
VISA-13B~\cite{yan2024visa}                & {\color[HTML]{656565} \tiny [ECCV\textquotesingle24]}                   &                                                                                                       & 52.3                     & 55.8                     & 54.1   & 38.3    & 43.5    & 40.9      & 45.3    & 49.7   & 47.5                             \\
VISA(IT)-7B~\cite{yan2024visa}             & {\color[HTML]{656565} \tiny [ECCV\textquotesingle24]}                   &                                                                                                       & 49.2                     & 52.6                     & 50.9   & 40.6    & 45.4    & 43.0      & 44.9    & 49.0   & 46.9                             \\
VISA(IT)-13B~\cite{yan2024visa}            & {\color[HTML]{656565} \tiny [ECCV\textquotesingle24]}                   &                                                                                                       & 55.6                     & 59.1                     & 57.4   & 42.0    & 46.7    & 44.3      & 48.8    & 52.9   & 50.9                             \\
VRS-HQ-7B~\cite{gong2025devil}               & {\color[HTML]{656565} \tiny [CVPR\textquotesingle25]}                   &                                                                                                       & \multicolumn{1}{l}{59.8} & \multicolumn{1}{l}{64.5} & 62.1   & 53.5    & 58.7    & 56.1      & 56.6    & 61.6   & 59.1          \\
GLUS~\cite{lin2025glus}                    & {\color[HTML]{656565} \tiny [CVPR\textquotesingle25]}                   &                                                                                                       & \multicolumn{1}{l}{58.3} & \multicolumn{1}{l}{56.0} & 60.7   & 48.8    & 53.9    & 51.4      & -       & -      & -              \\
HyperSeg~\cite{wei2025hyperseg}                & {\color[HTML]{656565} \tiny [CVPR\textquotesingle25]}                   &                                                                                                       & \multicolumn{1}{l}{56.0} & \multicolumn{1}{l}{60.9} & 58.5   & 50.2    & 55.8    & 53.0      & 53.1    & 58.4   & 55.7             \\
RGA3-3B~\cite{wang2025object}                 & {\color[HTML]{656565} \tiny [ICCV\textquotesingle25]}                   &            & 57.6 & 61.0 & 59.3   & 50.6    & 55.0    & 52.8      & 54.1    & 58.0   & 56.1          \\
RGA3-7B~\cite{wang2025object}                 & {\color[HTML]{656565} \tiny [ICCV\textquotesingle25]}                   &                                                                                                       & 58.7 & 62.3 & 60.5   & 53.1    & 57.7    & 55.4      & 55.9    & 60.0   & 58.0        \\
ViLLa~\cite{zheng2025villa}                   & {\color[HTML]{656565} \tiny [ICCV\textquotesingle25]}                   &                                                                                                       & -    & -    & -      & -       & -       & -         & 54.9    & 59.1   & 57.0           \\
InstructSeg~\cite{wei2025instructseg}             & {\color[HTML]{656565} \tiny [ICCV\textquotesingle25]}                   &                                                                                                       & 54.8 & 59.2 & 57.0   & 49.2    & 54.7    & 51.9      & 52.0    & 56.9   & 54.5           \\ \hline
CoT-RVS-online-7B~\cite{CoTRVS}                                            & {\color[HTML]{656565} \tiny [arXiv\textquotesingle25]} &  \multirow{2}{*}{\begin{tabular}[c]{@{}c@{}}VLM-Based\\ w/ Explicit \end{tabular}}  & -             & -             & -                          & -             & -             & -                          & 43.5          & 48.8          & 46.2                                                                  \\
CoT-RVS-offline-12B~\cite{CoTRVS}                                          & {\color[HTML]{656565} \tiny [arXiv\textquotesingle25]} &                                                                                     & -             & -             & -                          & -             & -             & -                          & 43.4          & 50.9          & 47.1                                                              \\
\rowcolor{gray!30}\mymodel-7B (Ours)                                                 & \multicolumn{1}{c}{\color[HTML]{656565} \tiny -}         & Reasoning & \textbf{63.3}          & \textbf{68.1}          & \textbf{65.7}                       & \textbf{55.4}          & \textbf{61.8}          & \textbf{58.6}                       & \textbf{59.3}          & \textbf{65.0}          & \textbf{62.1}                                                                 \\ \hline

\end{tabular}

 \vspace{-0.05in}
\end{table*}

\noindent\textbf{Training Datasets.} 
For our efficient RL post-training, we rely solely on the video object segmentation (VOS) data, in contrast to previous works~\cite{bai2024one,yan2024visa, gong2025devil,wei2025hyperseg,wei2025instructseg} that jointly fine-tunes on large, heterogeneous corpora spanning video segmentation, image segmentation, and VQA datasets.
Specifically, we curate training data from five benchmarks: Ref-YouTube-VOS~\cite{seo2020urvos}, MeViS~\cite{ding2023mevis}, Ref-DAVIS17~\cite{khoreva2018video}, ReVOS~\cite{yan2024visa} and LV-VIS~\cite{wang2023towards}. 
For each annotated sequence, we first convert per-frame masks to bounding boxes, which serve as ground-truth signals for post-training rewards. To ensure label quality, we run SAM2~\cite{ravi2024sam} on every frame conditioned on its ground-truth box to obtain predicted masks, compute IoU against the annotated masks, and discard all queries from any video whose mean IoU falls below 0.6. This filtering yields approximately 67k data pairs.

\noindent\textbf{Benchmarks.} We evaluate on five standard VOS benchmarks: two reasoning datasets including ReVOS~\cite{yan2024visa} and ReasonVOS~\cite{bai2024one} and three referring datasets including Ref-DAVIS17~\cite{khoreva2018video}, Ref-YouTube-VOS~\cite{seo2020urvos}, MeViS~\cite{ding2023mevis}. Notably, ReasonVOS has no training split, thus its evaluation is zero-shot, providing a clearer measure of the model's generalization ability.

\noindent\textbf{Baselines.} We benchmark against three families of methods to contextualize our gains. (1) \textit{Segmentation Specialists}: strong VOS/Ref-VOS systems~\cite{botach2022end,wu2022language,ding2023mevis,li2024llama,seo2020urvos,ding2022language,wu2023onlinerefer,he2024decoupling,miao2023spectrum,luo2023soc} trained with dense supervision, optimized for mask quality and temporal consistency. (2) \textit{VLM-Based with Latent Tokens Methods}: methods~\cite{bai2024one,yan2024visa,gong2025devil,lin2025glus,wei2025hyperseg,wei2025instructseg,zheng2025villa,wang2025object} that fine-tune VLMs to emit task-specific control tokens or logits that drive a downstream mask head, which are the current mainstream for reasoning VOS. (3) \textit{VLM-Based with Explicit Reasoning Methods}: methods that perform reasoning to explicitly ground targets via boxes/masks, an under-explored paradigm where CoT-RVS~\cite{CoTRVS} and our method fall. We report results across all three to isolate the advantage of our proposed \mymodel.

\noindent\textbf{Evaluation Metrics.} Following previous works~\cite{bai2024one,yan2024visa, gong2025devil,wei2025hyperseg,wei2025instructseg,zheng2025villa,lin2025glus,CoTRVS} on reasoning video object segmentation, we report region similarity ($\mathcal{J}$), contour accuracy ($\mathcal{F}$) and their mean ($\mathcal{J}\&\mathcal{F}$) as the primary video-level metrics.

\label{sec:exp settings}
\noindent\textbf{Implementation Details.} We adopt Qwen2.5-VL-7B~\cite{bai2025qwen2} as the default reasoning model $\mathcal{F}$ and SAM2 (Hiera-L)~\cite{ravi2024sam} as the default video tracker model $\mathcal{T}$. For post-training $\mathcal{F}$ with GRPO, each optimizer step processes 128 input data, and we sample $n=8$ rollouts per prompt, yielding an effective batch of 1024 sequences per optimizer step. 
The learning rate is set to $1e-6$, and the KL regularization coefficient $\beta=1e-3$.
For each video, we uniformly sample 16 frames as input to $\mathcal{F}$. All input frames are resized to $448\times 448$ before the first round generation. In the second round, the selected keyframe $I_k$ is resized to $840\times 840$ for spatial grounding. The tracker $\mathcal{T}$ operates on the full video at its original resolution.

\subsection{Experimental Results}
\label{sec:main results}

\noindent\textbf{\textit{Reasoning} Video Object Segmentation.} We first evaluate on reasoning VOS benchmarks -- ReasonVOS dataset in \cref{tab:reasonvos} and ReVOS dataset in \cref{tab:revos}. On ReasonVOS, \mymodel-7B achieves a decisive margin over the previous state-of-the-art (SOTA) method RGA-7B \cite{wang2025object}, improving $\mathcal{J}$ by +10.5 points, $\mathcal{F}$ by +11.7 points, and $\mathcal{J}\&\mathcal{F}$ by +11.2 points.
The significant performance improvement demonstrates the effectiveness of \mymodel\ with the proposed framework.
Furthermore, given the zero-shot nature of ReasonVOS, these gains highlight our strong generalization and robustness under challenging open-world queries.

On ReVOS, we conduct a comprehensive comparison across nine metrics. Our \mymodel-7B consistently ranks first, surpassing previous SOTAs, including several larger parameter systems, by a obvious margin. 
The across-the-board improvements on these reasoning VOS benchmarks substantiate the effectiveness of our explicit reasoning chain and the efficiency of the proposed training recipe.

\noindent\textbf{\textit{Referring} Video Object Segmentation.}
As  previous practice \cite{yan2024visa,bai2024one,wang2025object,zheng2025villa,ding2023mevis,khoreva2018video,seo2020urvos}, we report the experiment results on three Ref-VOS benchmarks, i.e., Ref-YouTube-VOS, Ref-DAVIS17, and MeViS. 
As  in \cref{tab:referring vos}, consistently, our \mymodel-7B sets a new state of the art, improving $\mathcal{J}\&\mathcal{F}$ by +2.7 points on Ref-YouTube-VOS, +4.8 points on Ref-DAVIS17, and +8.5 points on MeViS against the previous SOTAs. 
Notably, MeViS is a motion-guided benchmark and is regarded as
the most challenging Ref-VOS benchmark in GLUS~\cite{lin2025glus}. Our substantial gains on MeViS indicate strong adaptability on complex video scenarios with intricate motion patterns.
Although referring queries require less semantic reasoning than reasoning queries, our performance gains are still obvious and consistent. These results indicate \mymodel\ offers stronger cross-modal video understanding, better temporal aggregation, and more accurate target object detection than prior art.

\begin{table}[]\centering
\caption{Zero-shot reasoning image segmentation results on ReasonSeg~\cite{lai2024lisa} dataset.\label{tab:image grounding} }
\resizebox{0.8\columnwidth}{!}{
\begin{tabular}{ccccc}
\hline
\multirow{2}{*}{Method} & \multicolumn{2}{c}{test}        & \multicolumn{2}{c}{val} \\ \cline{2-5} 
                        & gIoU & \multicolumn{1}{c}{cIoU} & gIoU       & cIoU       \\ \hline
Qwen2.5VL-7B            & 55.9 & \multicolumn{1}{c}{44.3} & 59.5       & 54.0       \\
\mymodel-7B (Ours)                    & \textbf{59.7} & \multicolumn{1}{c}{\textbf{47.4}} & \textbf{63.7}       & \textbf{59.9}       \\ \hline
\end{tabular}}

\vspace{-0.1in}
\end{table}

\begin{table*}[]\footnotesize\centering
\caption{Video referring segmentation results on Ref-Youtube-VOS~\cite{seo2020urvos}, Ref-DAVIS17~\cite{khoreva2018video}, MeViS~\cite{ding2023mevis} datasets.}
\begin{tabular}{llcccccccccc}
\hline
\multicolumn{2}{c}{\multirow{2}{*}{Method}}  & \multicolumn{1}{c}{\multirow{2}{*}{Type}}                                                             & \multicolumn{3}{c}{Ref-YouTube-VOS}                        & \multicolumn{3}{c}{Ref-DAVIS17}                            & \multicolumn{3}{c}{MeViS}                                  \\ \cline{4-12} 
                         &                            & \multicolumn{1}{l}{}                                                                                  & $\mathcal{J}$ & $\mathcal{F}$ & $\mathcal{J}\&\mathcal{F}$ & $\mathcal{J}$ & $\mathcal{F}$ & $\mathcal{J}\&\mathcal{F}$ & $\mathcal{J}$ & $\mathcal{F}$ & $\mathcal{J}\&\mathcal{F}$ \\ \hline
URVOS~\cite{seo2020urvos}                    & {\color[HTML]{656565} \tiny [ECCV\textquotesingle20]}                   & \multirow{7}{*}{\begin{tabular}[c]{@{}c@{}}Segmentation\\ Specialists\end{tabular}}                   & 45.3          & 49.2          & 47.2                       & 47.3          & 56.0          & 51.6                       & 25.7          & 29.9          & 27.8                       \\
LBDT~\cite{ding2022language}                     & {\color[HTML]{656565} \tiny [CVPR\textquotesingle22]}                   &                                                                                                       & 48.2          & 50.6          & 49.4                       & -             & -             & 54.1                       & 27.8          & 30.8          & 29.3                       \\
MTTR~\cite{botach2022end}                     & {\color[HTML]{656565} \tiny [CVPR\textquotesingle22]}                   &                                                                                                       & 54.0          & 56.6          & 55.3                       & -             & -             & -                          & 28.8          & 31.2          & 30.0                       \\
LMPM~\cite{ding2023mevis}                     & {\color[HTML]{656565} \tiny [ICCV\textquotesingle23]}                   &                                                                                                       & -             & -             & -                          & -             & -             & -                          & 34.2          & 40.2          & 37.2                       \\
ReferFormer~\cite{wu2022language}              & {\color[HTML]{656565} \tiny [CVPR\textquotesingle22]}                   &                                                                                                       & 61.3          & 64.6          & 62.9                       & 58.1          & 64.1          & 61.1                       & 29.8          & 32.2          & 31.0                       \\
OnlineRefer~\cite{wu2023onlinerefer}              & {\color[HTML]{656565} \tiny [CVPR\textquotesingle23]}                   &                                                                                                       & 61.6          & 65.5          & 63.5                       & 61.6          & 67.7          & 64.8                       & -             & -             & -                          \\
DsHmp~\cite{he2024decoupling}                    & {\color[HTML]{656565} \tiny [CVPR\textquotesingle24]}                   &                                                                                                       & 65.0          & 69.1          & 67.1                       & 61.7          & 68.1          & 64.9                       & 43.0          & 49.8          & 46.4                       \\ \hline
LISA-7B~\cite{lai2024lisa}                  & {\color[HTML]{656565} \tiny [CVPR\textquotesingle24]}                   & \multirow{13}{*}{\begin{tabular}[c]{@{}c@{}}VLM-Based\\ w/ Latent\\Tokens\end{tabular}}   & 53.4          & 54.3          & 53.9                       & 62.2          & 67.3          & 64.8                       & 35.1          & 39.4          & 37.2                       \\
LISA-13B~\cite{lai2024lisa}                 & {\color[HTML]{656565} \tiny [CVPR\textquotesingle24]}                   &                                                                                                       & 54.0          & 54.8          & 54.5                       & 63.2          & 68.8          & 66.9                       & 35.8          & 40.0          & 37.9                       \\
TrackGPT-7B~\cite{zhu2023tracking}              & {\color[HTML]{656565} \tiny [arXiv\textquotesingle24]}                  &                                                                                                       & 55.3          & 57.4          & 56.4                       & 59.4          & 67.0          & 63.2                       & 37.6          & 42.6          & 40.1                       \\
TrackGPT-13B~\cite{zhu2023tracking}             & {\color[HTML]{656565} \tiny [arXiv\textquotesingle24]}                  &                                                                                                       & 58.1          & 60.8          & 59.5                       & 62.7          & 70.4          & 66.5                       & 39.2          & 43.1          & 41.2                       \\ 
VISA-7B~\cite{yan2024visa}                  & {\color[HTML]{656565} \tiny [ECCV\textquotesingle24]}                   &   & 59.8          & 63.2          & 61.5                       & 66.3          & 72.5          & 69.4                       & 40.7          & 46.3          & 43.5                       \\
VISA-13B~\cite{yan2024visa}                 & {\color[HTML]{656565} \tiny [ECCV\textquotesingle24]}                   &                                                                                                       & 61.4          & 64.7          & 63.0                       & 67.0          & 73.8          & 70.4                       & 41.8          & 47.1          & 44.5                       \\
VideoLISA~\cite{bai2024one}                & {\color[HTML]{656565} \tiny [NeurIPS\textquotesingle24]}                &                                                                                                       & 61.7          & 65.7          & 63.7                       & 64.9          & 72.7          & 68.8                       & 41.3          & 47.6          & 44.4                       \\
VRS-HQ-7B~\cite{gong2025devil}                & {\color[HTML]{656565} \tiny [CVPR\textquotesingle25]}                   &                                                                                                       & 68.3          & 72.5          & 70.4                       & 72.6          & 79.4          & 76.0                       & 47.6          & 53.7          & 50.6                       \\
GLUS~\cite{lin2025glus}                     & {\color[HTML]{656565} \tiny [CVPR\textquotesingle25]}                   &                                                                                                       & 65.5          & 69.0          & 67.3                       & -             & -             & -                          & 48.5          & 54.2          & 51.3                       \\
HyperSeg~\cite{wei2025hyperseg}                 & {\color[HTML]{656565} \tiny [CVPR\textquotesingle25]}                   &                                                                                                       & -             & -             & 68.5                       & -             & -             & 71.2                       & -             & -             & -                          \\
RGA3-3B~\cite{wang2025object}                  & {\color[HTML]{656565} \tiny [ICCV\textquotesingle25]}                   &                                                                                                       & 65.8          & 69.1          & 67.4                       & 67.6          & 76.6          & 72.1                       & 46.2          & 51.5          & 48.8                       \\
RGA3-7B~\cite{wang2025object}                  & {\color[HTML]{656565} \tiny [ICCV\textquotesingle25]}                   &                                                                                                       & 66.8          & 70.1          & 68.5                       & 68.3          & 77.3          & 72.8                       & 47.4          & 52.8          & 50.1                       \\
ViLLa~\cite{zheng2025villa}                    & {\color[HTML]{656565} \tiny [ICCV\textquotesingle25]}                   &                                                                                                       & 64.6          & 70.4          & 67.5                       & 70.6          & 78.0          & 74.3                       & 46.5          & 52.3          & 49.4                       \\
InstructSeg~\cite{wei2025instructseg}              & {\color[HTML]{656565} \tiny [ICCV\textquotesingle25]}                   &                                                                                                       & 65.4          & 69.5          & 67.5                       & 67.3          & 74.9          & 71.1                       & -             & -             & -                          \\ \hline
CoT-RVS-online-7B~\cite{CoTRVS}                                           & {\color[HTML]{656565} \tiny [arXiv\textquotesingle25]}   & \multirow{2}{*}{\begin{tabular}[c]{@{}c@{}}VLM-Based \\w/ Explicit \end{tabular}}  & -             & -             & -                          & 70.4          & 77.5          & 73.9                       & 42.7          & 49.1          & 45.9                       \\
CoT-RVS-offline-13B~\cite{CoTRVS}                                         & {\color[HTML]{656565} \tiny [arXiv\textquotesingle25]}   &                                                                                   & -             & -             & -                          & 70.9          & 78.3          & 74.6                       & 40.3          & 48.1          & 44.2                       \\
\rowcolor{gray!30}\mymodel-7B    (Ours)                                           & \multicolumn{1}{c}{\color[HTML]{656565} \tiny -}           &  Reasoning & \textbf{71.1}          & \textbf{75.2}          & \textbf{73.1}                       & \textbf{77.4}          & \textbf{84.1}          & \textbf{80.8}                       & \textbf{56.1}          & \textbf{63.4}          & \textbf{59.8}                       \\ \hline 
\end{tabular}
\label{tab:referring vos}
\end{table*}

\noindent\textbf{Zero-shot Reasoning \textit{Image} Segmentation.}
Besides the main results, we ask whether post-training on \textit{video} tasks improves spatial grounding that transfers to \textit{image} reasoning segmentation. In other words, does our model truly learn better spatial grounding capabilities which is generalizable to other tasks?
To probe this, we conduct a zero-shot evaluation on the reasoning image segmentation task.
Specifically, we evaluate both the base model (Qwen2.5-VL-7B~\cite{bai2025qwen2}) and our post-training \mymodel-7B on ReasonSeg dataset~\cite{lai2024lisa}, and report the mean per-image IoU (gIoU) and the cumulative IoU (cIoU) as in LISA~\cite{lai2024lisa}.
As shown in \cref{tab:image grounding}, despite no image-specific training, \mymodel\ yields delivers consistent gains on both test and validation splits compared to the base model. These improvements indicate that our pipeline no only simply couples spatial grounding with video understanding and temporal cues, but also upgrades its intrinsic spatial grounding ability.

\begin{figure*}
    \centering
    \includegraphics[width=0.95\linewidth]{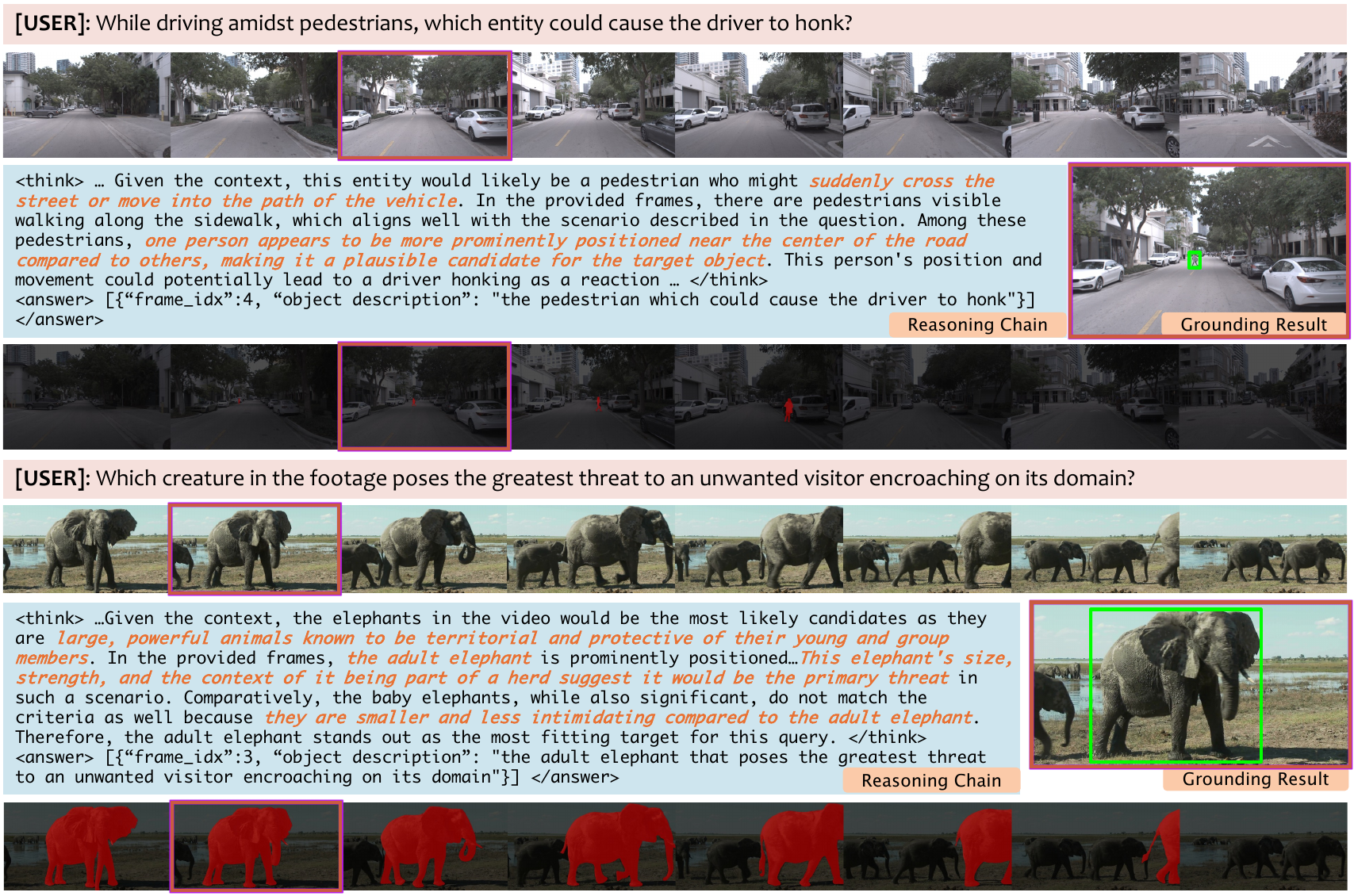}
        \vspace{-0.1in}
    \caption{Qualitative cases of \mymodel\ on ReasonVOS~\cite{bai2024one}. The frame highlighted in red indicates the selected keyframe. The green bounding box within the enlarged keyframe on the right size represents the grounding result. Zoom in to view visual details. \label{fig:quanlitative results} }
   
     \vspace{-0.05in}
\end{figure*}

\noindent \textbf{Qualitative Results}.
\cref{fig:quanlitative results} presents detailed case studies highlighting the role of explicit reasoning chains in VOS. In the first example, the video depicts a typical road scene without a visually salient target. The query asks what could triggered the driver’s honk. The model parses the scene, integrates visual cues with traffic commonsense, and infers plausible causes, ultimately identifying the jaywalking pedestrian as the most likely trigger. Notably, the target object occupies only a small portion of the frame, making this case particularly challenging. In the second example, the model is queried about the creature posing the greatest threat. It first recognizes the most menacing species (elephant) and, guided by world knowledge that elder elephants protect calves and the herd, pinpoints the specific individual with high precision. Across cases, the reasoning chain consistently selects well-segmentable keyframes, stabilizing grounding and mitigating error accumulation, which yields cleaner masks and more consistent trajectories.
Additional visualizations are provided in the supplementary.

\subsection{Ablation Studies}
\label{sec:ablation studies}
\begin{table}[]
\centering\footnotesize
\caption{Ablation experiments for the proposed decoupling framework and RL post-training. The referring VOS results were evaluated on the Ref-DAVIS17 dataset, while the reasoning VOS results were evaluated on the ReasonVOS dataset. Decom. Pipeline indicates whether to use the decomposed pipeline, RL denotes the usage of Reinforcement Learning.}
\begin{tabular}{cccccccc}
\hline
\multirow{2}{*}{\begin{tabular}[c]{@{}c@{}}Decom.\\ Pipeline\end{tabular}} & \multirow{2}{*}{\begin{tabular}[c]{@{}c@{}}RL\end{tabular}} & \multicolumn{3}{c}{Ref-DAVIS17}                                            & \multicolumn{3}{c}{ReasonVOS}                                            \\ \cline{3-8}   &   & $\mathcal{J}$        & $\mathcal{F}$        & $\mathcal{J}\&\mathcal{F}$ & $\mathcal{J}$        & $\mathcal{F}$        & $\mathcal{J}\&\mathcal{F}$ \\ \hline
 &   &   3.2&   5.6&    4.4& 7.4& 8.8&  8.1  \\   & \checkmark                                                         & 4.0                &   6.0     &       5.3     &               7.1  &    8.1     &   7.6            \\ 
\checkmark                          &                                                                       &59.2&               66.9&               63.0&                       38.7&                 43.4&        41.0        \\
\checkmark                          & \checkmark                                                         & 77.4           & 84.1               &  80.7                &  61.8             &  67.7              & 64.8                   \\ \hline 
\end{tabular}
\label{tab:decompose and rl}
\vspace{-0.05in}
\end{table}

\noindent\textbf{Key Pipeline Design.}
We isolate the contributions of decomposed reasoning chain and RL post-training via four variants. (1) \textit{Base Model}, which directly predicts spatio-temporal grounding from the video and query. (2)  \textit{Base Model + RL}, which applies GRPO post-training to the base model. (3) \textit{Decomposed}, which restructures inference into our multi-turn decomposed reasoning chain. (4) \textit{Decomposed + RL}, our final model \mymodel\ with GRPO-enhanced reasoning. We evaluate them on Ref-DAVIS17 (in-domain Ref-VOS) and ReasonVOS (out-of-domain reasoning VOS), the results are showed in \cref{tab:decompose and rl}. 
The base model exhibits very poor video grounding ability, and RL alone fails to lift performance due to sparse effective rollouts and credit assignment in end‑to‑end prediction. In contrast, decomposed pipeline yields a clear jump by guiding the model to compose its primitive skills (video understanding, temporal grounding, spatial grounding) into a coherent procedure. Adding RL further drives self-evolution, tightening the interplay of these skills for VOS reasoning and delivering substantial additional gains across both datasets. Together, decomposed reasoning and RL post-training are necessary and complementary, culminating in unprecedented performance.

\begin{table}[]
\caption{Results obtained by employing different frame sampling rates during both the training and testing phases separately. \label{tab:frame num} }

\resizebox{\columnwidth}{!}{
\begin{tabular}{cccccc}
\hline
\multirow{2}{*}{\begin{tabular}[c]{@{}c@{}}\#Training \\ Frames\end{tabular}} & \multirow{2}{*}{\begin{tabular}[c]{@{}c@{}}\#Testing \\ Frames\end{tabular}} & \multicolumn{3}{c}{Ref-DAVIS17}                              & \multirow{2}{*}{\begin{tabular}[c]{@{}c@{}}Training Time\\ per Step (s)\end{tabular}} \\ \cline{3-5} &  & $\mathcal{J}$ & $\mathcal{F}$ & $\mathcal{J}\&\mathcal{F}$ &  \\ \hline
12    & 12  & 76.8          & 83.8          & 80.3                       & 603      \\
16   & 16                                                                           & \textbf{77.4}          & \textbf{84.1}          & \textbf{80.7}                       & 725          \\
20   & 20  &          77.0 &          \textbf{84.1} &   80.5                    & 831   \\ \hline 
16   & 12  & 76.6          & 83.9          & 80.3                       & \multirow{3}{*}{725}   \\
16   & 16  & 77.4          & \textbf{84.1}          & 80.7                       &    \\
16   & 20  & \textbf{77.5}          & \textbf{84.1}          & \textbf{80.8}                       &              \\ \hline
\end{tabular}}

\vspace{-0.05in}
\end{table}

\noindent\textbf{Frame Sampling Strategy.}
We examine the effect of input frame count during training and inference. As shown in \cref{tab:frame num}, we train models with 12 / 16 / 20 uniformly sampled frames and observe the diminishing returns beyond 16. Accuracy improves from 12 to 16, while additional frames bring only marginal gains but increase training cost roughly. Balancing efficiency and performance, we adopt 16 frames as the default for both training and evaluation. At test time, performance remains stable across different frame counts, indicating that the reasoning pipeline is robust to temporal sampling density.

\begin{table}[]\small
\caption{Model performance on VOS with three different temporal reward types. The referring VOS results were evaluated on the MeViS dataset, while the reasoning VOS results were evaluated on the ReasonVOS dataset. \label{tab:temporal reward} }
\vspace{-0.05in}
\begin{tabular}{ccccccc}
\hline
\multirow{2}{*}{\begin{tabular}[c]{@{}c@{}}Temporal\\ Reward Type\end{tabular}} & \multicolumn{3}{c}{MeViS}                              & \multicolumn{3}{c}{ReasonVOS}                              \\ \cline{2-7}& $\mathcal{J}$ & $\mathcal{F}$ & $\mathcal{J}\&\mathcal{F}$ & $\mathcal{J}$ & $\mathcal{F}$ & $\mathcal{J}\&\mathcal{F}$ \\ \hline
No Reward                                                                         &50.7         & 58.7         &   54.7                   & 55.0        & 61.0         & 58.0                      \\
0/1 Reward                                                                        &53.6          &61.2          & 57.4                     & 56.7        & 62.7         & 59.7                      \\
Soft Reward                                                                       &\textbf{56.1}          &\textbf{63.4}          & \textbf{59.8}                       & \textbf{61.8}          & \textbf{67.7}         & \textbf{64.8}                      \\ \hline
\end{tabular}
\vspace{-0.1in}
\end{table}

\noindent\textbf{Designs of Reward.}
\label{sec:temporal rewards}
Temporal rewards are crucial for teaching the model to select keyframes that genuinely facilitate downstream spatial grounding and mask decoding. Prior keyframe selection works~\cite{wu2019adaframe,liang2024keyvideollm,zhang2025q,hu2025m} primarily score semantic relevance, which is insufficient for VOS: a good frame should also reveal the target with minimal occlusion and adequate scale. We therefore ablate three temporal reward schemes in RL: (1) no temporal reward, (2) a binary 0/1 reward that only checks whether the target appears in the selected frame(s), and (3) our soft temporal reward that scores object visibility via normalized bounding box area. As shown in \cref{tab:temporal reward}, the soft variant provides a graded learning signal aligned with object visibility and scale, alleviates sparse credit assignment, and discourages degenerate selections (e.g., heavily occluded or tiny targets).

\section{Conclusion}
\label{sec:conclusion}

In this work, we reformulate reasoning-centric VOS as a sequential decision problem and introduce \mymodel, which decomposes the task into three pretrained primitive capabilities. We further develop a reinforcement learning framework that directly optimizes the reasoning chain, enabling the model to refine its decision quality without relying on dense supervision. Experiments demonstrate state-of-the-art performance across multiple VOS benchmarks and uncover interpretable, stepwise reasoning trajectories. We believe that this explicit-chain, outcome-driven formulation provides a general paradigm for advancing reasoning-aligned video understanding models.

{
    \small
    \bibliographystyle{ieeenat_fullname}
    \bibliography{main}
}

\appendix
\clearpage
\appendix
\onecolumn

\section*{Appendix}

\section{Training Curves}

\begin{figure*}[h]
	\centering
 	\subfloat[]{\includegraphics[width=.3\linewidth]{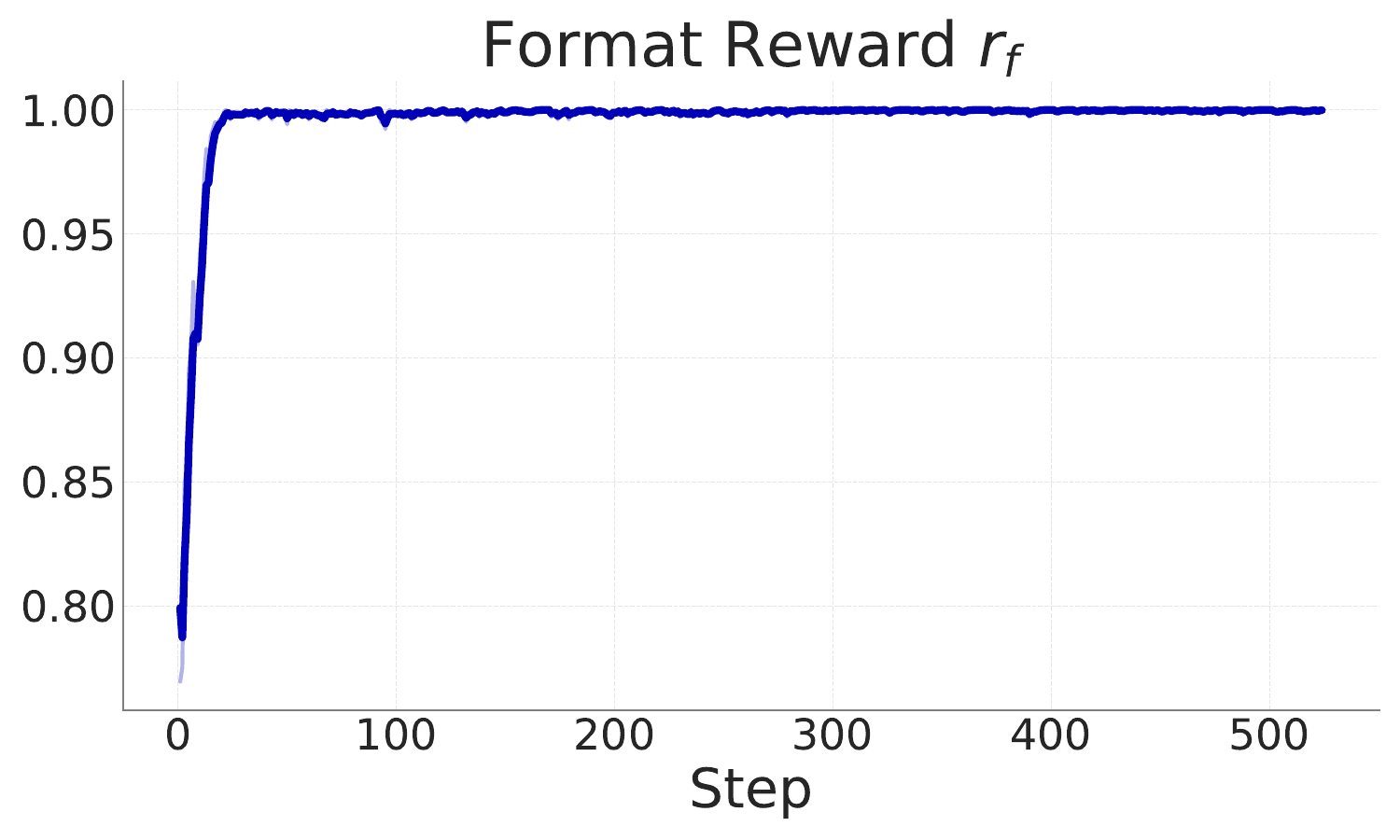}}\hspace{5pt}
	\subfloat[]{\includegraphics[width=.3\linewidth]{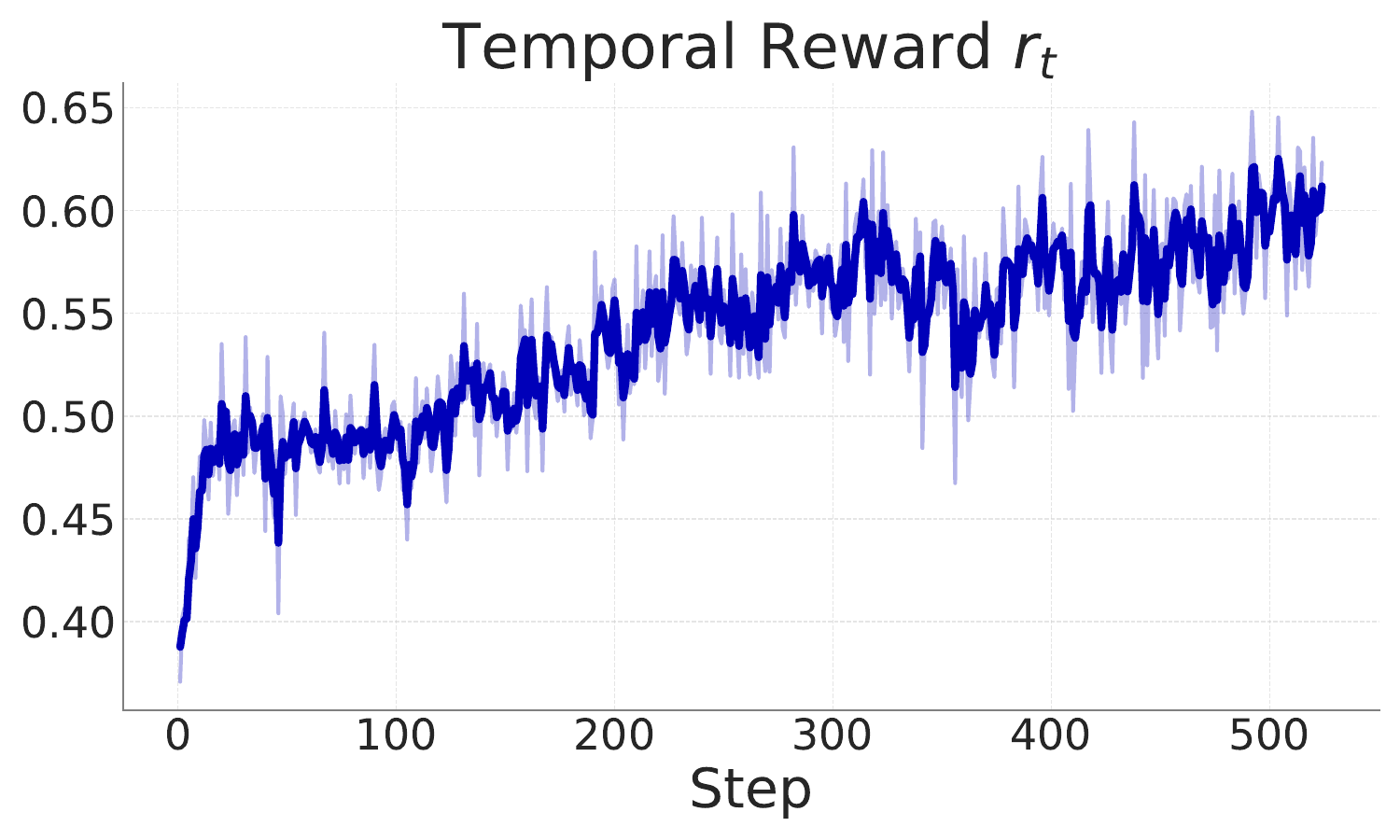}}\hspace{5pt}
	\subfloat[]{\includegraphics[width=.3\linewidth]{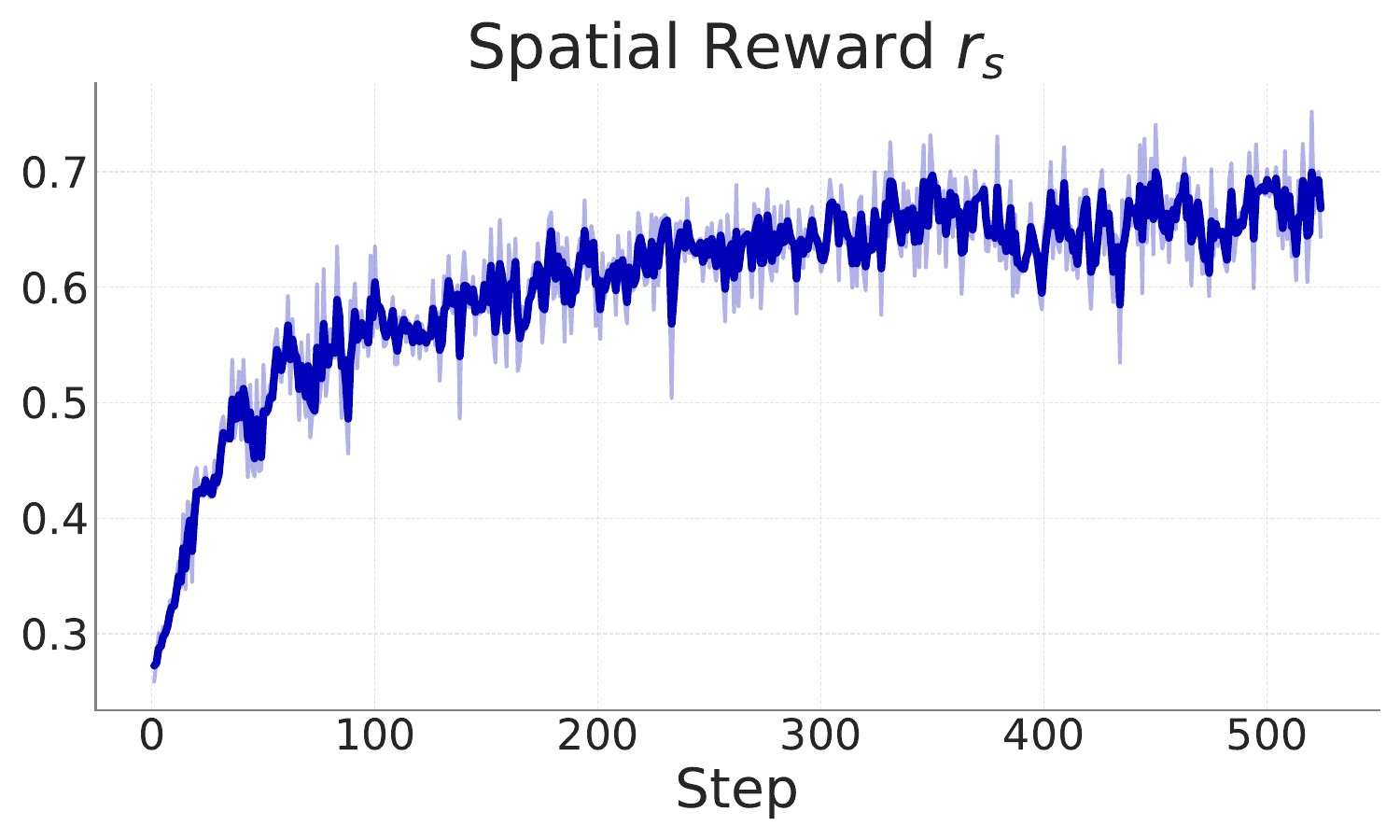}}\\
	\subfloat[]{\includegraphics[width=.3\linewidth]{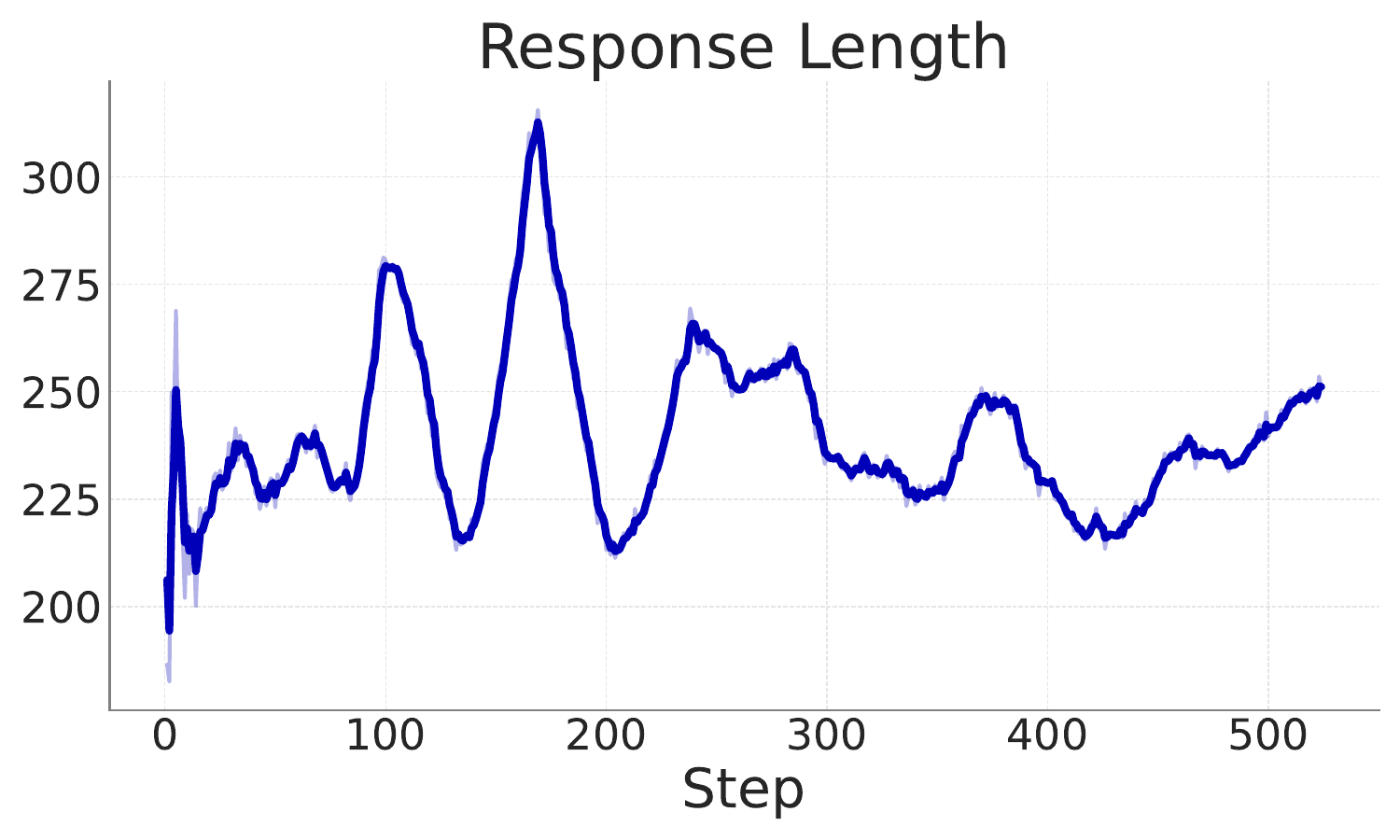}}\hspace{5pt}
        \subfloat[]{\includegraphics[width=.3\linewidth]{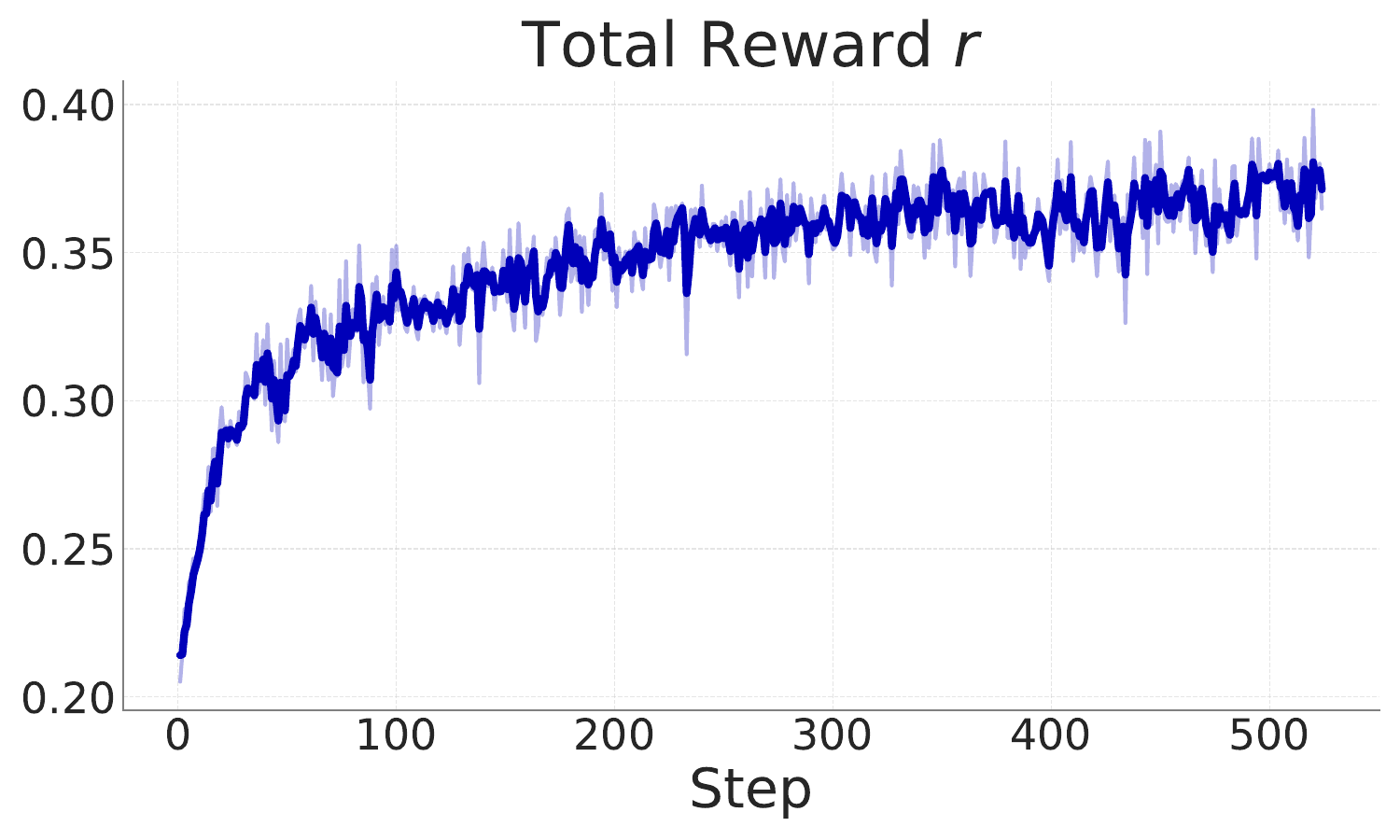}}\hspace{5pt}
	\subfloat[]{\includegraphics[width=.3\linewidth]{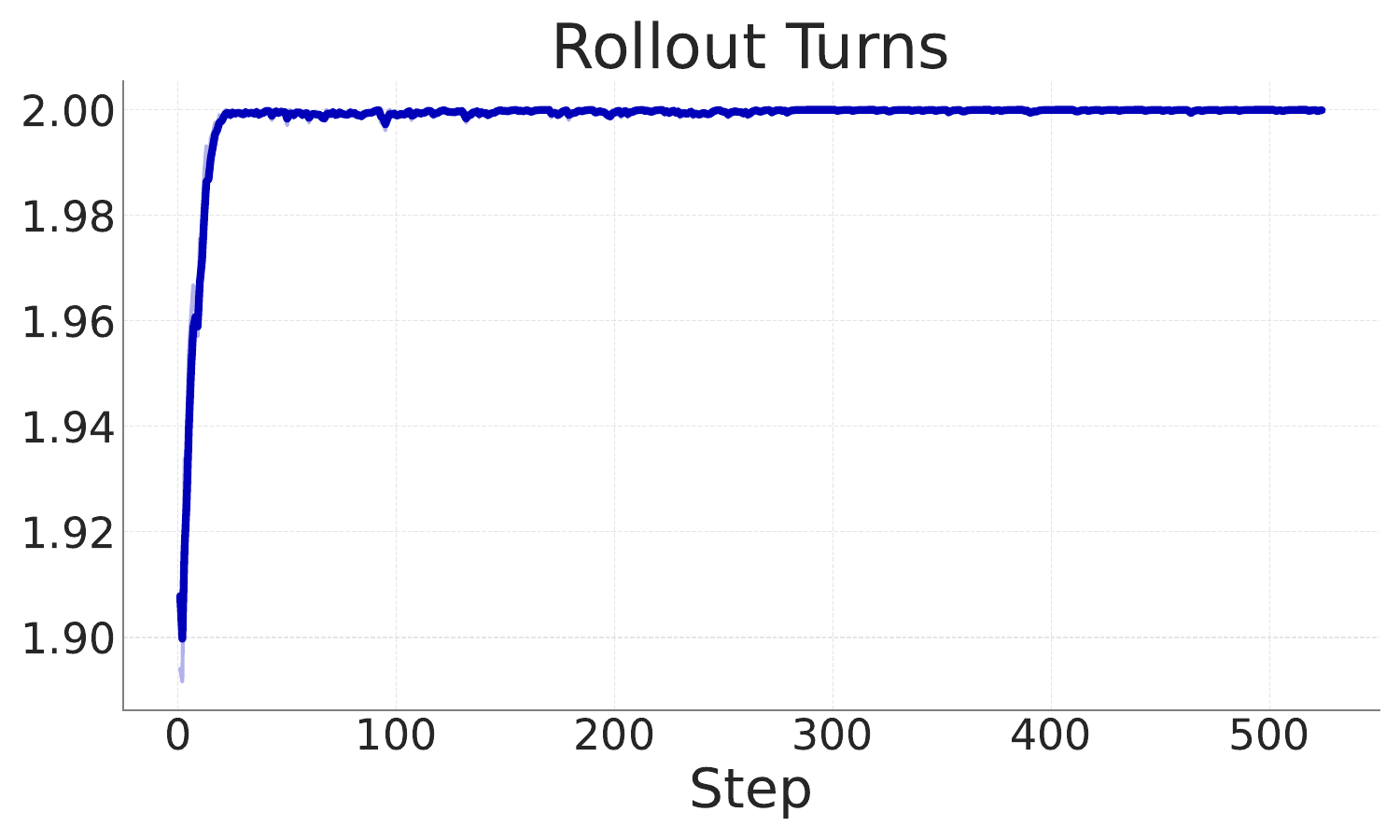}}\hspace{5pt}
	\caption{Training curves of \mymodel. (a) Format reward $r_f$ rapidly converges to a full score and remains saturated. (b) Temporal reward $r_t$ and (c) Spatial reward $r_s$ increase steadily with training. (d) Response length remains stable overall without collapse. (e) Total reward $r$ rises consistently over time. (f) Average number of rollout turns quickly converge to 2. \label{fig: suppl training curve}}
\end{figure*}

In \cref{fig: suppl training curve}, we summarize the dynamics of \mymodel\ training. 
The first row plots the three reward components. Format reward $r_f$ in (a) rises sharply to near-perfect score within the first few steps and remains saturated thereafter, indicating that the policy quickly masters the instructed output format and consistently completes the two-round rollout generations. Temporal reward $r_t$ and spatial reward $r_s$ in (b) and (c) follow a similar growth pattern early on, suggesting that initial gains are driven primarily by producing well-formed, parseable responses that expose the temporal-spatial grounding results. As the format reward plateaus, the growth of temporal and spatial rewards decelerates improvements, steming from stronger reasoning rather than formatting, reflecting better temporal selection and tighter localization.

The second row reports response length, total reward $r$ and rollout turns. Response length in (d) increases during the early phase as the proportion of complete two round rollouts rises, and shows pronounced oscillations between roughly 100 and 250 steps. The analysis of training output reveals several shifts in response policy before settling into to a stable reasoning pattern. Total reward in (e) exhibits a steady increase over training. Rollout turns in (f), defined as the average number of rounds per sample within a batch, quickly converge to two, consistent with the observed trends of the reward components.

Overall, these curves indicate a well-behaved optimization, \mymodel\ promptly adopts the intend two-round policy and continues to improve its reasoning quality in a stable manner.

\begin{figure*}
    \centering
    \includegraphics[width=0.95\linewidth]{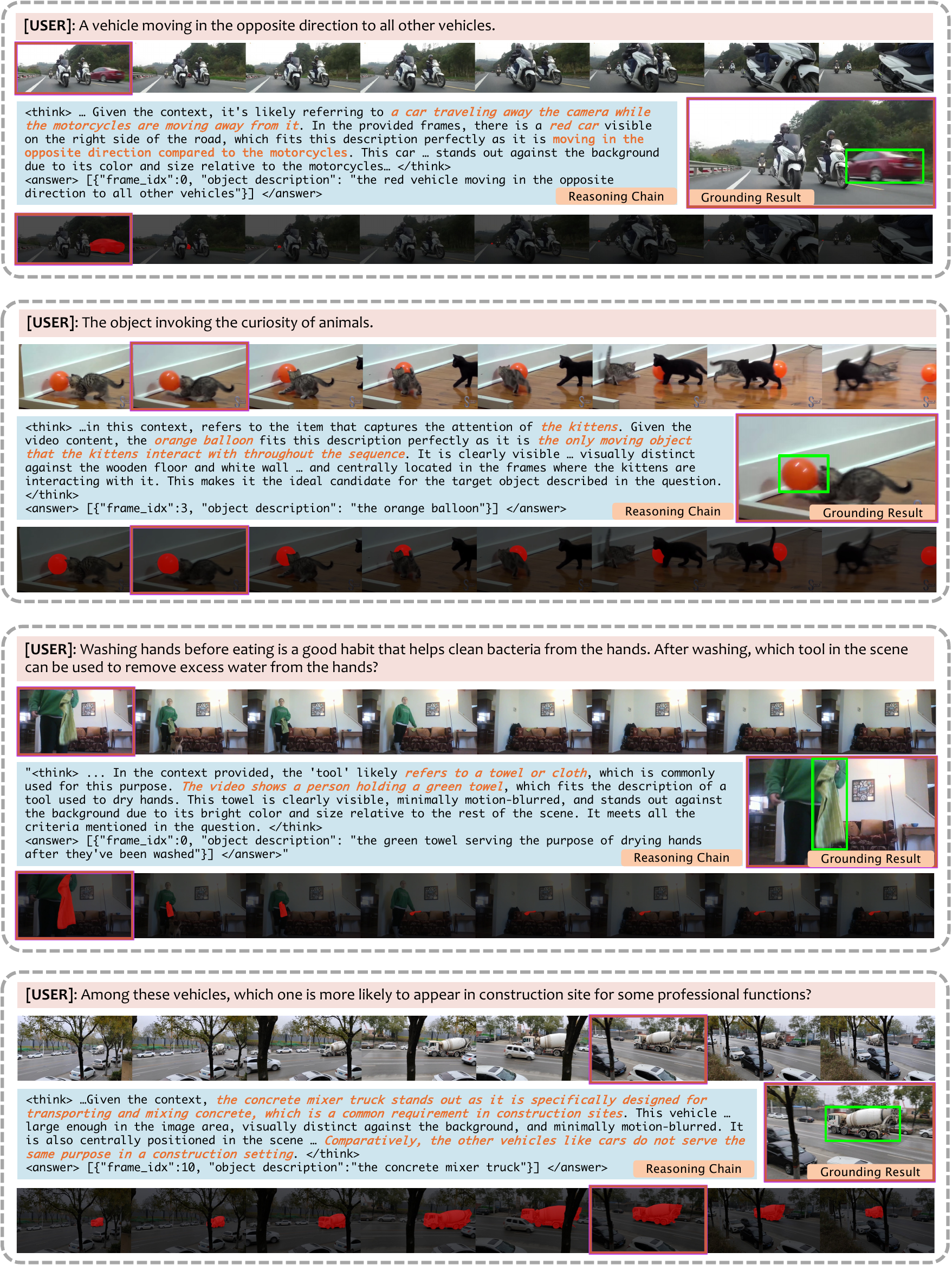}
    \caption{Additional qualitative cases of \mymodel. The frame highlighted in red indicates the selected keyframe. The green bounding box within the enlarged keyframe on the right side represents the grounding result. Zoom in to view visual details. \label{fig:suppl quanlitative results} }
   
\end{figure*}

\section{Qualitative Results}

In \cref{fig:suppl quanlitative results}, we present more visualization examples highlighting the reasoning-centric segmentation capability of \mymodel. Across diverse scenes, the model analyzes temporal dynamics, including object interactions (\eg, the kittens interact with orange ballon, person holding green towel) and motion cues (\eg, car moving in the opposite direction compared to the motorcycles), while aligning them with the abstract semantics of the user query. It integrates the video evidence with commonsense knowledge (\eg, the concrete mixer truck designed for construction sites) to identify the target entity, selects frames that are favorable for downstream localization (\eg, bright color, centrally located in the frames, visually against the background, size), and converts this selection into a keyframe index and a concrete object description. Conditioned on this description, \mymodel\ produces tight spatial grounding and propagates masks across the video sequence, yielding the fine-grained segmentation masks.

\section{Implementation Details}
In this section, we supplement some implementation details, including user prompts in \cref{sec: suppl prompt} and pseudocode for our proposed decomposed generation with reasoning chain in \cref{sec: suppl alg}.
\subsection{\mymodel\ User Prompt Details}\label{sec: suppl prompt}
\begin{figure*}
  \centering
  \subfloat[User prompt for first round generation, where $\{second\},\{nframes\}$ and $\{question\}$ refer to the video duration $T/fps$, the frame number $T$ of input video sequence $V$ and input query $\mathbf{x}$, respectively.]
  {\includegraphics[width=0.9\textwidth]{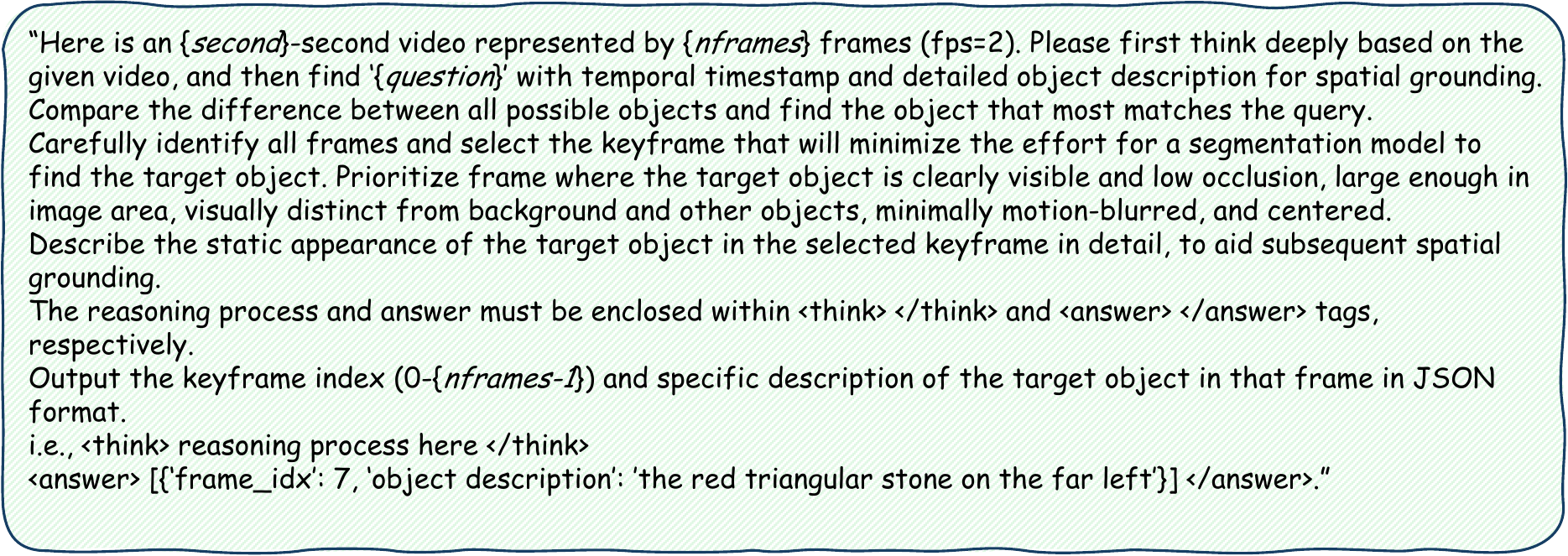}}\quad
  \subfloat[User prompt for second round generation, where $\{object\ description\}$ refers to the extracted object description $\mathbf{d}$ from the first round response $\mathbf{y_1}$.]
  {\includegraphics[width=0.9\textwidth]{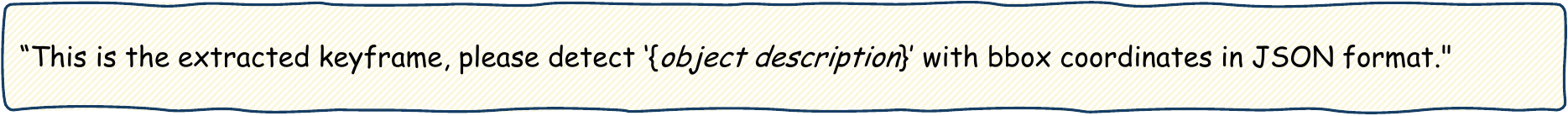}}
  \caption{User prompt templates for two-round rollout.\label{fig:prompt}}
\end{figure*}

To elicit reliable reasoning for video object segmentation, we design a task-specific prompting scheme tailored to our two-round rollout of the policy VLM $\mathcal{F}$, as shown in \cref{fig:prompt}. Each round uses a distinct prompt that mirrors the decomposition of the task.

\noindent\textbf{Round one: user prompt for video understanding and temporal grounding.} The first prompt follows a chain-of-thought style to encourage deep analysis. It asks the model to compare salient objects across the video, examine their occurrences over frames, and select a keyframe where the target is most suitable for localization. The prompt instructs $\mathcal{F}$ to first output a reasoning trace, then a structured result containing the keyframe index and a concise object description.

\noindent\textbf{Round two: user prompt for spatial grounding.} The second prompt provides the keyframe and object description extracted from round one response and requests precise localization on the selected frame. The output must follow a JSON format containing a tight bounding box.

Both prompts request reasoning separated from the final answer, standardized JSON fields to stabilize generation and reduce parsing errors. In practice, this design yields interpretable responses, that serve as robust visual prompts for subsequent mask propagation.

\subsection{Algorithm}\label{sec: suppl alg}
We distill the key generation component of \mymodel\ into a compact pseudocode summarized in \cref{alg:vos}. For each input video-query input, the model performs $n$ times two-round rollout sampling and ultimately returns the resulting trajectories set $\{o_i\}_{i=1}^n$.

\begin{algorithm*}[]\small
\caption{Decomposed Generation of \mymodel\ with Two-Round Rollout \label{alg:vos}}
\DontPrintSemicolon
\SetKwInOut{KwIn}{Require}
\SetKwInOut{KwOut}{Ensure}
\KwIn{Video $V=\{I_t\}_{t=1}^T$, query $\mathbf{x}$, policy VLM $\mathcal{F}$, parser $\mathcal{G}$, group size $n$, prompt template $\mathcal{P}_1$ and $\mathcal{P}_2$}
\KwOut{Rollout sequences $\{o_i\}_{i=1}^n$}

\BlankLine
\For{$i=1$ \KwTo $n$}{
Initialize current VLM input sequence $\mathbf{z}\leftarrow \mathcal{P}_1(V,\mathbf{x})$\;
Initialize current VLM rollout sequence $o_i\leftarrow\varnothing$\;
\BlankLine
\tcp{Round-1: video understanding + temporal grounding}
Generate the first round response sequence $\mathbf{y_1} \sim \mathcal{F}(\cdot \mid \mathbf{z})$\;
Extract status flag $S_1\in\{\texttt{succ,fail}\}$, keyframe index $k$ and object description $\mathbf{d}$ through parser $(S_1, k, \mathbf{d}) \leftarrow \mathcal{G}(\mathbf{y_1})$\;
Append $\mathbf{y}_1$ to input sequence $\mathbf{z}\leftarrow \mathbf{z} \oplus \mathbf{y}_1$\;
Append $\mathbf{y}_1$ to rollout sequence $o_i\leftarrow o_i \oplus \mathbf{y}_1$\;
\If{$S_1=\texttt{fail}$}{
    Terminate rollout generation early and flag status $S_2\leftarrow\texttt{fail}$, set keyframe $I_k$ and bounding box $B_k$ to Null $I_k, B_k\leftarrow \texttt{null}$\;
    \textbf{Continue}
}
\BlankLine
\tcp{Round-2: spatial grounding}
Select the keyframe $I_k$ from video $I_k\leftarrow V[k]$\;
Append $(I_k, \mathbf{d})$ to the ongoing input sequence $\mathbf{z}\leftarrow \mathbf{z}\oplus \mathcal{P}_2(I_k,\mathbf{d})$\;
Generate the second response sequence $\mathbf{y_2} \sim \mathcal{F}(\cdot \mid \mathbf{z})$\;
Extract status flag $S_2\in\{\texttt{succ,fail}\}$ and spatial grounding result $B_k\in\mathbb{R}^4_{\geq 0}$ through parser $(S_2,B_k) \leftarrow \mathcal{G}(\mathbf{y_2})$\;
Append $\mathbf{y_2}$ to rollout sequence $o_i \leftarrow o_i \oplus \mathbf{y_2}$
}
\Return{final generated rollouts $\{o_i\}_{i=1}^n$}
\end{algorithm*}

\end{document}